\documentclass[lettersize,journal]{IEEEtran}
\usepackage{amsmath,amsfonts}
\usepackage{ulem}
\usepackage{array}
\usepackage[caption=false,font=normalsize,labelfont=sf,textfont=sf]{subfig}
\usepackage{textcomp}
\usepackage{stfloats}
\usepackage{url}
\usepackage{verbatim}
\usepackage{graphicx}
\usepackage{cite}
\usepackage{mathrsfs}
\usepackage{multirow}
\usepackage{makecell}
\usepackage{xcolor}
\usepackage{booktabs}
\usepackage{hyperref}
\usepackage{bbding}
\usepackage{pifont}
\usepackage{algorithm}%
\usepackage{algorithmicx}%
\usepackage{algpseudocode}%
\usepackage{listings}%

\begin{document}

\title{MID: A Self-supervised Multimodal Iterative Denoising Framework}

\author{Chang Nie, Tianchen Deng, Zhe Liu,~\IEEEmembership{Member,~IEEE,} and Hesheng Wang,~\IEEEmembership{Senior Member,~IEEE,}
\thanks{This work was supported in part by the Natural Science Foundation of China under Grant 62225309, U24A20278, 62361166632 and U21A20480. (Corresponding Author: Hesheng Wang, e-mail: wanghesheng@sjtu.edu.cn)}%
\thanks{Chang Nie, Tianchen Deng, Zhe Liu and Hesheng Wang are with School of Automation and Intelligent Sensing, Shanghai Jiao Tong University and Key Laboratory of System Control and Information Processing, Ministry of Education of China, Shanghai 200240, China.}
}

\markboth{Journal of \LaTeX\ Class Files,~Vol.~14, No.~8, August~2021}%
{Shell \MakeLowercase{\textit{et al.}}: A Sample Article Using IEEEtran.cls for IEEE Journals}


\maketitle

\begin{abstract}
  Data denoising is a persistent challenge across scientific and engineering domains. Real-world data is frequently corrupted by complex, non-linear noise, rendering traditional rule-based denoising methods inadequate. To overcome these obstacles, we propose a novel self-supervised multimodal iterative denoising (MID) framework. MID models the collected noisy data as a state within a continuous process of non-linear noise accumulation. By iteratively introducing further noise, MID learns two neural networks: one to estimate the current noise step and another to predict and subtract the corresponding noise increment. For complex non-linear contamination, MID employs a first-order Taylor expansion to locally linearize the noise process, enabling effective iterative removal. Crucially, MID does not require paired clean–noisy datasets, as it learns noise characteristics directly from the noisy inputs. Experiments across four classic computer vision tasks demonstrate MID's robustness, adaptability, and consistent state-of-the-art performance. Moreover, MID exhibits strong performance and adaptability in tasks within the biomedical and bioinformatics domains.
\end{abstract}


\section{Introduction}
\label{sec:intro}
\IEEEPARstart{H}{igh}-quality data is foundational for reliable scientific discovery and technological innovation. However, noise contamination during real-world acquisition is nearly ubiquitous, arising from sensor limitations, transmission errors, environmental interference, or intrinsic system stochasticity. Such contamination not only obscures the underlying signal but also degrades the performance of downstream algorithms in domains such as computer vision, medical diagnosis, and biological signal interpretation. Consequently, effective denoising has become an indispensable preprocessing step across a wide range of fields.

\begin{figure}[t]
  \centering
   \includegraphics[width=0.99\linewidth]{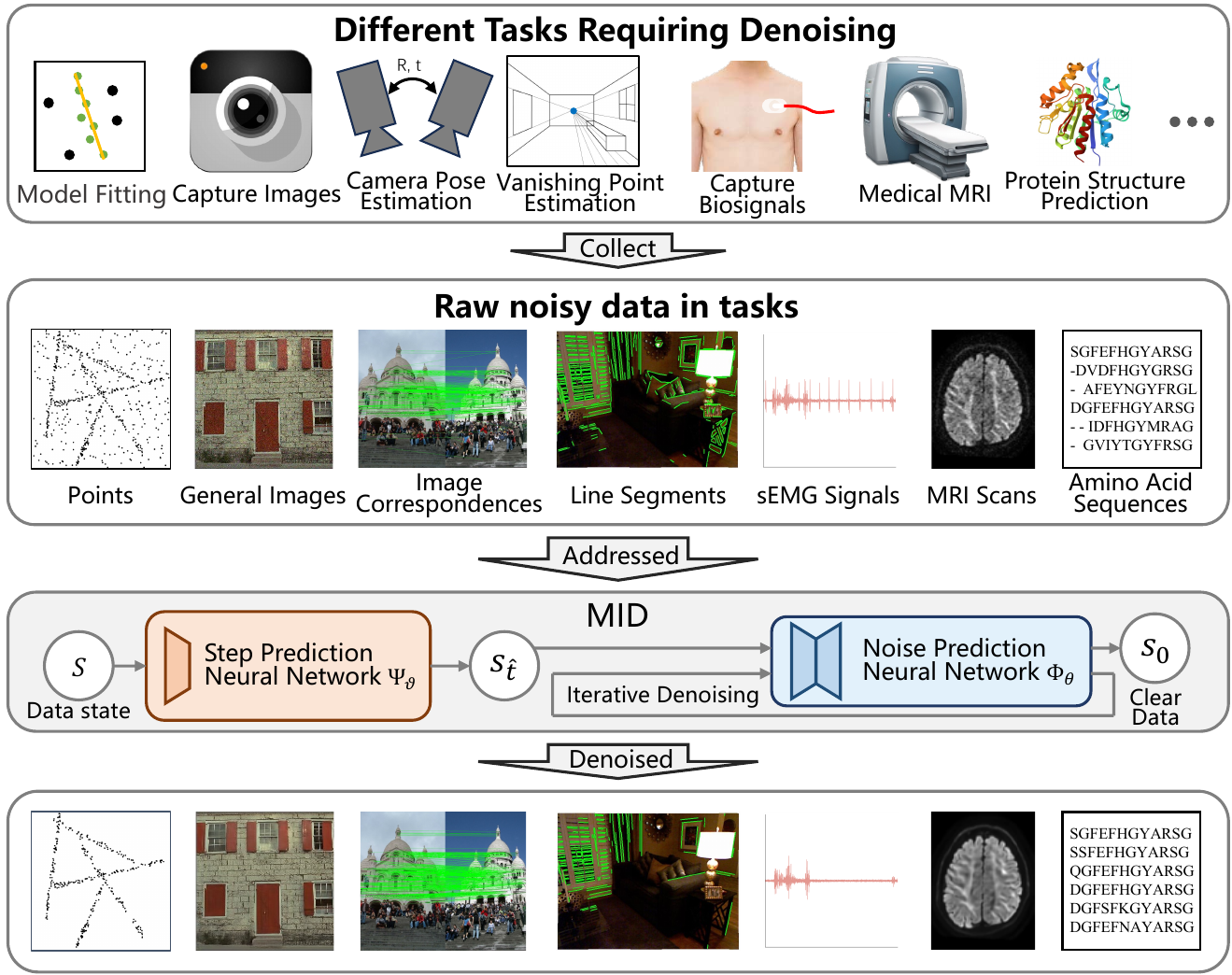}
    \vspace{-0.4cm}
   \caption{\textbf{Overview of the MID denoising framework.} MID processes raw noisy data from diverse modalities and domains. The framework first estimates the noise severity and then executes iterative denoising steps to progressively restore the clean data.
   }
   \label{fig:headpic}
\end{figure}

Existing denoising methods possess notable limitations. Traditional approaches, including model-based filtering, transform-domain thresholding, and non-local means, are often tailored to specific noise models or data types \cite{gupta2023image, pfaff2023self, yang2023chaotic}. While they may achieve strong performance under their assumed conditions, their efficacy often diminishes when noise statistics deviate from those assumptions. In recent years, supervised deep learning has achieved impressive results but typically demands large paired datasets of clean and noisy samples \cite{fang2020multilevel, ma2025pixel2pixel, chen2019real}. Such data are often costly or infeasible to collect in practice, particularly in fields like clinical imaging or remote sensing where clean ground truth is unavailable. Self-supervised learning alleviates the need for paired data by exploiting data redundancies or structural priors \cite{fadnavis2020patch2self, lehtinen2018noise2noise, li2024stimulating, wang2025m3dm}. However, these methods can suffer from oversmoothing, loss of fine detail, or an over-reliance on modality-specific assumptions, such as spatial correlation in images.

To address these challenges, we propose the MID framework, a self-supervised Multimodal Iterative Denoising method. As illustrated in Fig. \ref{fig:headpic}, the self-supervised MID first estimates the noise level of the data using a step prediction neural network $\Psi_\vartheta$. It then employs a noise prediction neural network $\Phi_\theta$ to predict and eliminate noise within the current data state through an iterative denoising loop. This process gradually restores clean data, which can significantly improve the performance of downstream tasks.

This paper details principles of the MID framework and demonstrates its efficacy across a wide spectrum of applications. These include reducing model fitting outliers, denoising images, removing incorrect correspondences, optimizing line segments, purifying biological signals like sEMG, enhancing medical image quality in MRI, and improving the utility of bioinformatics data such as amino acid sequences. Our findings suggest that MID offers a robust, adaptable, and broadly applicable solution to the persistent challenge of denoising.

The main contributions of MID are as follows:
\begin{itemize}
	\item We propose MID, a novel Multimodal Iterative Denoising framework. Its self-supervised training paradigm enables effective denoising on multimodal data from diverse tasks without requiring clean ground-truth data.

	\item We introduce a mechanism that learns to recognize and remove noise by progressively exposing the system to further corruption. This process utilizes two neural networks: one estimates the noise level, and the other predicts and subtracts the noise based on that estimate. We innovatively employ a first-order Taylor expansion to linearize complex non-linear noise contamination processes, enabling their reversal through iterative subtraction.

	\item MID demonstrates exceptional cross-domain applicability in computer vision, bioinformatics, medicine, and biology, achieving significant performance improvements.
    
\end{itemize}

\section{Related Work}
\label{sec:Related_Work}
We review existing work in four representative application domains where denoising plays a critical role: image denoising, biological signal denoising, MRI denoising, and denoising of amino acid sequences. For each domain, we outline typical approaches, their strengths and limitations, and how they motivate the design of the proposed MID framework.

\subsection{Image Denoising}
Effective image denoising is crucial for a wide range of computer vision applications, as noise can significantly degrade visual information. Traditional methods, which often rely on manually designed functions and patch-based filtering, can lack robustness in diverse, real-world scenarios \cite{luo2015adaptive}. Supervised learning methods, while powerful, require extensive paired clean and noisy image datasets, which are often difficult and expensive to obtain \cite{zhou2020supervised, chaudhary2022fast}. To overcome this limitation, self-supervised methods have been developed to train denoisers using only noisy images. Among these, Blind2Unblind \cite{wang2022blind2unblind} introduces a re-visible loss to transition from blind-spot to non-blind denoising, thereby preventing information loss. To better handle large-scale, spatially correlated noise, MM-BSN \cite{zhang2023mm} proposes a multi-mask strategy using variously shaped kernels to break noise correlations more effectively. Another approach, C-BSN \cite{jang2023self}, uses a downsampled invariance loss and a conditional blind-spot network to selectively use center pixel information while a random subsampler decorrelates noise. Theoretical frameworks like Noisier2Noise have also been extended to justify and improve existing methods by introducing corrective loss weightings \cite{millard2023theoretical}. However, these self-supervised techniques can still risk losing image details or over-smoothing results. MID improves on these approaches by learning noise characteristics directly from noisy inputs, applying a Taylor expansion to handle non-linear contamination, and refining results iteratively to preserve fine details.

\subsection{Biological Signal Denoising}
Biological signals, such as surface electromyography (sEMG), are widely used for clinical diagnosis \cite{vijayvargiya2024s, li2024human, wei2024continuous}, rehabilitation monitoring, and human–machine interaction. In practice, these signals are frequently contaminated by electrocardiogram (ECG) interference, which overlaps in frequency content and can severely affect feature extraction. Several methods have been developed to address this challenge. While high-pass filtering is a straightforward technique \cite{drake2006elimination, ince2009ego}, it often removes important low-frequency components of the sEMG signal along with the ECG interference. Template subtraction methods \cite{drake2006elimination, ince2009ego} work by identifying a repeating ECG pattern and subtracting it from the recording; however, these methods rely on rigid assumptions about signal distributions, limiting their robustness. More advanced neural network-based methods have shown significant promise but can be unstable during training and may introduce unwanted distortions or artifacts \cite{chiang2019noise, fan2020vibration, cui2022gan, wang2022ecg}. A robust solution must adapt to variations in both signals while avoiding the loss of clinically relevant features. MID addresses this by iteratively modeling and subtracting structured noise without relying on paired clean–noisy data, enabling stable learning even under complex interference conditions.

\subsection{MRI Denoising}
Magnetic Resonance Imaging (MRI) is a non-invasive technique that provides high-resolution anatomical and functional information. However, clinical demands for faster scans often lead to lower signal-to-noise ratios (SNR), necessitating advanced denoising strategies to preserve diagnostic quality. Early self-supervised methods such as Noise2Noise \cite{lehtinen2018noise2noise} train a network to map one noisy realization of an image to another, but can oversmooth fine anatomical structures. Patch2Self \cite{fadnavis2020patch2self} leverages redundancy across multiple 3D volumes in diffusion MRI, but it requires training a separate model per volume, which risks spatial inconsistency. More recently, DDM$^2$ \cite{xiangddm} integrated statistical denoising with generative diffusion models, treating the noisy input as an intermediate diffusion state. While this approach achieves strong results, it can be sensitive to initial noise levels and modality-specific tuning. These limitations highlight the need for a more general, self-supervised, and detail-preserving approach.

\begin{figure*}[t]
  \centering
   \includegraphics[width=0.90\linewidth]{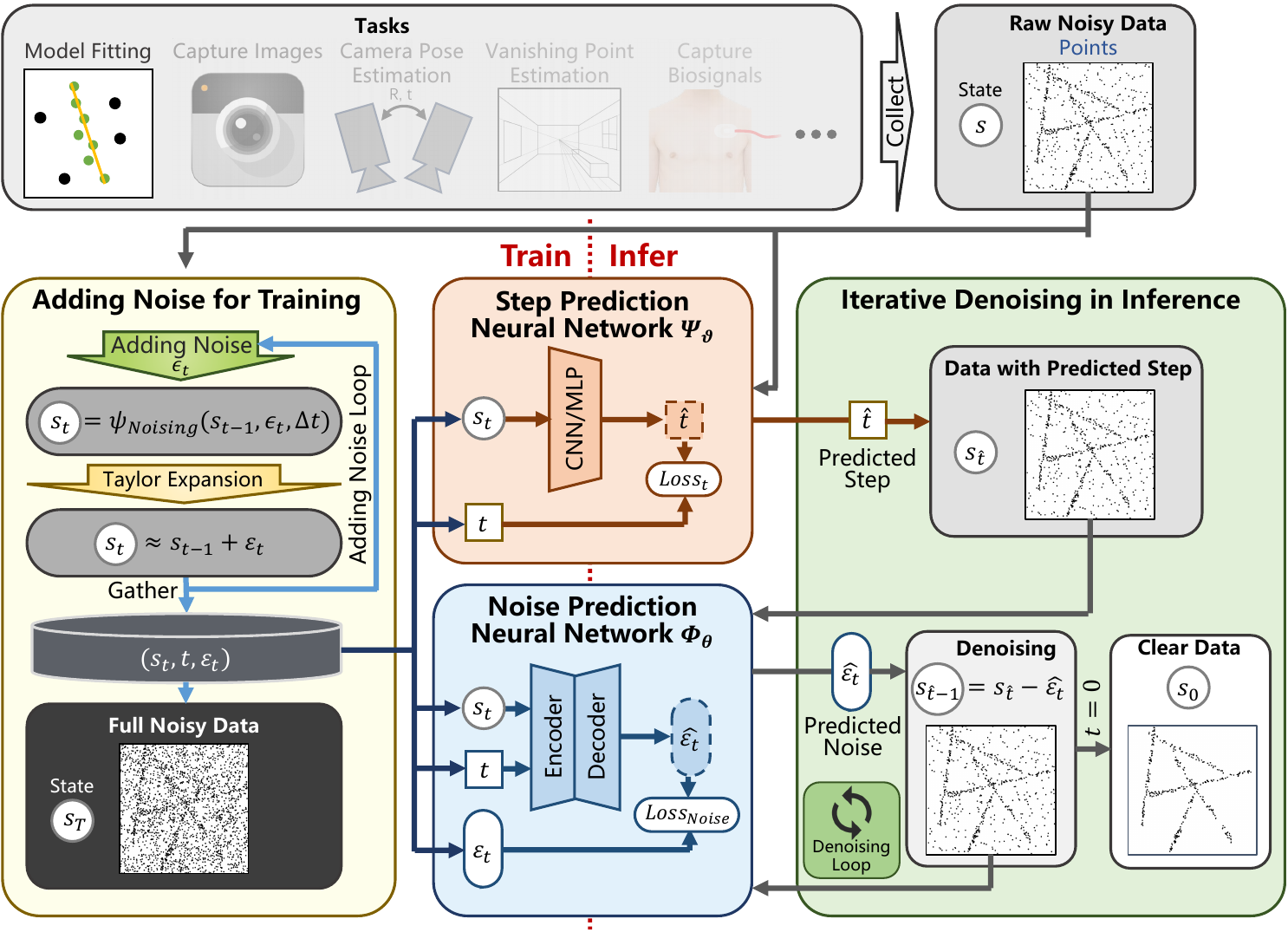}
   \caption{\textbf{The MID training and denoising pipeline, illustrated with a model-fitting outlier task.} The denoising of outliers in the model fitting task serves as an example. (a)  During self-supervised training, raw data is treated as an initial state in a continuous, non-linear noise addition process. By linearizing this process with a Taylor expansion, MID learns to recognize noise steps and features by repeatedly adding noise. (b) For denoising, the step prediction network estimates the noise step of input. The reversible linearized process is then used to iteratively predict and subtract noise, effectively denoising the data.}
   \label{fig:overview}
\end{figure*}

\subsection{Denoising Amino Acid Sequences}
In bioinformatics, Multiple Sequence Alignments (MSAs) are essential for tasks like protein contact prediction and structure modeling. Large MSAs often contain redundant or misaligned sequences that introduce noise, increasing computational cost and potentially degrading prediction accuracy. To handle the massive size of MSAs, various subsampling and filtering methods are employed. A notable example is the greedy selection algorithm used by the MSA Transformer \cite{rao2021msa}, which iteratively builds a smaller MSA by selecting new sequences that maximize diversity based on Hamming distance. However, this reliance on a single criterion may not fully capture sequence quality or its value in downstream tasks. This limitation highlights the need for more advanced denoising and selection strategies. By reframing MSA optimization as a denoising task, MID can learn to distinguish informative sequences from noisy ones, thereby improving downstream prediction quality.

\begin{table*}[ht]\footnotesize 
\centering
\caption{\textbf{Comparison with existing denoising methods.}}
\setlength{\tabcolsep}{2.0mm}
\renewcommand\arraystretch{1.0}
\begin{tabular}{l|cccc}
\toprule
Methods           & Self-supervised & Denoising Image & Denoising Signal & Denoising Point Cloud \\ \hline\hline
DDM$^2$ \cite{xiangddm}             & \ding{51} & \ding{51} & \ding{55} & \ding{55} \\
Noise2Noise \cite{lehtinen2018noise2noise}     & \ding{51} & \ding{51} & \ding{55} & \ding{55} \\
Blind2Unblind \cite{wang2022blind2unblind}    & \ding{51} & \ding{51} & \ding{55} & \ding{55} \\
Neighbor2Neighbor \cite{huang2021neighbor2neighbor} & \ding{51} & \ding{51} & \ding{55} & \ding{55} \\
SDEMG \cite{liu2024sdemg}            & \ding{55} & \ding{55} & \ding{51} & \ding{55} \\
Ours              & \ding{51} & \ding{51} & \ding{51} & \ding{51} \\ \bottomrule
\end{tabular}
\label{tab:Comparison}
\end{table*}

The MID framework has several key attributes. It operates in a self-supervised manner, learning to denoise using only noisy data as input. This is achieved by further corrupting already noisy data and learning to reverse these incremental additions, which eliminates the need for pristine ground-truth samples. The framework employs an iterative process to subtract estimated noise, gradually recovering the clean signal while preserving fine details. Designed for generality, MID can be applied to diverse data types—including images, signals, and point cloud-like structures—without requiring fundamental architectural changes. Finally, its application of a Taylor expansion allows it to approximate complex non-linear noise contamination as a sequence of linear perturbations, enabling effective iterative removal.

\section{Methodology}
\subsection{Pipeline of MID}\label{principle}

MID trains a system to recognize and remove noise by learning from data, denoted by $s$, that is progressively corrupted. The framework, depicted in Fig. \ref{fig:overview}, involves two main phases: a training phase where noise characteristics are learned, and a denoising phase where the trained networks are used to remove noise from new, unseen data.

\subsection{Adding Noise for Training}\label{adding}
During training, MID takes an input data sample $s$, which is assumed to be noisy from real-world collection. Treating $s$ as an initial state $s_0$, MID simulates a noise accumulation process over $T$ steps. This recursive process is expressed as:
\vspace{-0.2cm}
\begin{equation}
s_t=\psi_{Noising}\left ( s, \epsilon_t ,t \right ), t= 1,2,...,T ,
\label{eq:noising}
\vspace{-0.1cm}
\end{equation}
where $s_t$ represents the data state at noise step $t$, $\epsilon_t$ is the noise added at that step, and $\psi_{Noising}$ describes the noise addition function. For linear noise, where noise combines additively ($s_{noisy}=s_{clean}+noise$) or multiplicatively which can be transformed to additive in the log domain ($s_{noisy}=s_{clean}\cdot noise$) with the signal, the process is expressed recursively:
\vspace{-0.2cm}
\begin{equation}
s_t=\psi_{Noising}\left ( s_{t-1}, \epsilon_{t} ,\Delta t \right ) .
\label{eq:recursive}
\vspace{-0.1cm}
\end{equation}

A key challenge arises when $\psi_{Noising}$ is non-linear, as is common with many real-world noise types (e.g., Poisson noise, JPEG artifacts, or structured noise like incorrect line segments). Reversing such a process directly is difficult. MID addresses this by approximating the non-linear addition at each step using a first-order Taylor expansion:
\vspace{-0.2cm}
\begin{equation}
s_t\approx s_{t-1}+\Delta s_{t-1}.
\label{eq:taylor}
\end{equation}
Here, the change $\Delta s_{t-1}$ is interpreted as an effectively additive noise term $\epsilon_{t}$ for that small step. This linearization transforms the complex problem of reversing non-linear noise addition into a sequence of simpler, linear noise subtraction steps. Higher-order terms are considered negligible as the noise increment $\epsilon_{t}$ at each step is kept small. Furthermore, in the approximation in Eq. \ref{eq:taylor}, if $s_t=\psi(s_{t-1},\epsilon_{t})$, then $s_t \approx \psi(\mathbf{s}_{t-1}, 0) + \left. \frac{\partial \psi}{\partial \epsilon_t} \right|_{\epsilon_t=0} \cdot \epsilon_t$. If $s_{t-1}$ is the state without the $k-th$ noise increment and $\psi(\mathbf{s}_{t-1}, 0)=s_{t-1}$, then $s_t \approx s_{t-1}+J \cdot \epsilon_t$. Our approach simplifies this to $s_t\approx s_{t-1}+{\epsilon_t}^\prime$, where ${\epsilon_t}^\prime$ is treated as the ``effective" additive noise that the Noise Prediction Network learns to predict. This implicitly means the Jacobian $J$ is either assumed to be close to identity or its effect is absorbed into the learned mapping by $\Phi_\theta$, which predicts ${\epsilon_t}^\prime$ directly from $s_t$ and $t$. For instance, if $s_{t-1}$ is the current image and Poisson noise $\epsilon_t$ is added, the noisy image $s_t \sim Poisson(s_{t-1} + \epsilon)$, which is non-linear. Our approach adds a controlled amount of noise $\epsilon$ such that $s_t=f(s_{t-1},\epsilon)$, and the linearization approximates this function $f$.

\begin{algorithm}[htb]
\small
\caption{Noise Addition Training Process}
\label{alg:train}
\begin{algorithmic} 
    \Require\\
        Noise steps $T$;\\
        Collected noisy data as state $s_0=s$;\\
        Initial step prediction network $\Psi_\vartheta$;\\
        Initial noise prediction network $\Phi_\theta$.
    \Ensure\\
        Trained step prediction network $\Psi_\vartheta$;\\
        Trained noise prediction network $\Phi_\theta$.\\
    
    \Repeat
        \For{$t=1$ to $T$}
            \State Sample noise $\epsilon_{t}$
            \State $s_t \gets \psi_{Noising}\left ( s_{t-1}, \epsilon_{t} \right )$
            \State {Taylor expansion: $s_t\approx s_{t-1}+\epsilon_t$}
            \State Take gradient descent step on $\nabla_\vartheta \left \| \Psi_\vartheta(s_t) - t \right \|$
            \State Take gradient descent step on $\nabla_\theta \left \| \Phi_\theta(s_t, t) - \epsilon_t \right \|$
        \EndFor
    \Until{convergence}
    \State \Return $\Psi_\vartheta$, $\Phi_\theta$
\end{algorithmic}
\end{algorithm}

With this linearized approximation, MID trains two neural networks in a self-supervised manner, as described in Algorithm \ref{alg:train}. The architectures of these networks are introduced in Section \ref{Architectures}. By repeatedly exposing itself to these controlled noise additions and learning to predict both the noise stage and the noise increment, MID builds a robust internal representation of noise characteristics.

\subsection{Iterative Denoising in Inference}\label{adding}

\begin{algorithm}[htb]
\small
\caption{Iterative Denoising Process}
\label{alg:sampling}
\begin{algorithmic} 
    \Require\\
        Collected noisy data as state $s$;\\
        Trained step prediction network $\Psi_\vartheta$;\\
        Trained noise prediction network $\Phi_\theta$.
    \Ensure\\
        Denoised data $s_0$.\\

    \State $\hat{t} \gets \Psi_\vartheta(s)$ \Comment{Predict the total noise steps from the data}
    \State $s_{\hat{t}} \gets s$ \Comment{Initialize the state at the max step}
    \For{$t \gets \hat{t}$ down to $1$}
        \State $s_{t-1} \gets s_t - \Phi_\theta(s_t, t)$ \Comment{Denoise by subtracting predicted noise}
    \EndFor
    \State \Return $s_0$
\end{algorithmic}
\end{algorithm}

Once trained, the MID framework can denoise new, unseen noisy data. As shown in Algorithm \ref{alg:sampling}, given a noisy input $s$, the Step Prediction Network $\Psi_\vartheta$ first estimates its current noise step, $\hat{t} = \Psi_\vartheta(s)$. The input data is then treated as the starting state $s_{\hat{t}}$. Subsequently, the Noise Prediction Network $\Phi_\theta$ predicts the noise component $\hat{\varepsilon}_t = \Phi_\theta(s_{\hat{t}}, \hat{t})$. This predicted noise is then subtracted to yield an improved estimate: $s_{\hat{t}-1} = s_{\hat{t}} - \hat{\varepsilon}_t$. This process is iterated—decrementing $t$ and repeating the noise prediction and subtraction steps—until $t$ reaches 0, yielding the final denoised data $s_0$. This iterative subtraction, guided by the learned noise features and the current noise level estimate, allows MID to progressively strip away layers of contamination.



\subsection{Network and Loss}\label{Architectures}

\textbf{Step Prediction Neural Network ($\Psi_\vartheta$):} This network learns to estimate the current noisy step $\hat{t}$ of a given noisy sample $s_t$. It is trained to minimize a loss function $Loss_t$ between the true step $t$ and the predicted step $\hat{t}$:
\begin{equation}
 \begin{aligned} & \underset{\vartheta}{\text{minimize}} & & Loss_{t}(t,\hat{t}) \\ & \text{s.t.} & & \hat{t}=\Psi_\vartheta(s_t) . \end{aligned}
\label{eq:pred_t}
\vspace{-0.1cm}
\end{equation}
For calculating the noise step prediction loss $Loss_t$, the true noise step $t\in \left [ 0,T \right ] $ is normalized to $\left [ 0,1 \right ] $. The Mean Squared Error (MSE) loss is used for this regression task:
\vspace{-0.2cm}
\begin{equation}
Loss_{t} = (t - \hat{t})^2 ,
\vspace{-0.1cm}
\end{equation}
where $\hat{t}$ is the prediction of the network.

\textbf{Noise Prediction Neural Network ($\Phi_\theta$):} This network learns to predict the additive noise component $\hat{\epsilon_t}$ introduced to reach state $s_t$ from $s_{t-1}$, given $s_t$ and the estimated step $\hat{t}$. It is trained to minimize a loss function $Loss_{Noise}$ between the actual added noise $\varepsilon_{t}=s_t-s_{t-1}$ (from the linearized approximation) and the predicted noise $\hat{\epsilon_t}$:
\begin{equation}
 \begin{aligned} & \underset{\theta}{\text{minimize}} & & Loss_{Noise}(\varepsilon_{t},\hat{\varepsilon}_{t}) \\ & \text{s.t.} & & \hat{\varepsilon}_{t}=\Phi_\theta(s_{\hat{t}}, t) . \end{aligned}
\label{eq:pred_noise}
\vspace{-0.1cm}
\end{equation}

For image-type data and signal data, the network $\Phi_\theta$ is trained with the noise prediction loss $Loss_{Noise}$ by minimizing the mean squared error (MSE) between the predicted noise $\hat{\varepsilon}_{t-1}$ and the actual added noise $\varepsilon_{t-1}$:
\vspace{-0.2cm}
\begin{equation}
Loss_{Noise} = \frac{1}{n} \sum_{i=1}^{n} (\varepsilon^i_{t} - \hat{\varepsilon}^i_{t})^2 ,
\vspace{-0.1cm}
\end{equation}
where $n$ is the number of data points.

For multidimensional point cloud data where denoising is a binary classification (identifying noisy points/sequences), we use the binary cross-entropy (BCE) loss to train this network:
\vspace{-0.2cm}
\begin{equation}
Loss_{Noise} = - \frac{1}{N} \sum_{i=1}^{N} \left[ y_i \log(p_i ) + (1 - y_i) \log(1 - p_i)  \right] ,
\vspace{-0.1cm}
\end{equation}
where $y_i$ and $p_i$ are the ground truth category and probability of a point being noise.
A multi-task loss combines the noise prediction loss and the noise step prediction loss:
\vspace{-0.2cm}
\begin{equation}
Loss_{total} = Loss_{Noise} + Loss_{t} .
\vspace{-0.1cm}
\end{equation}

\begin{figure*}[t]
  \centering
   \includegraphics[width=0.99\linewidth]{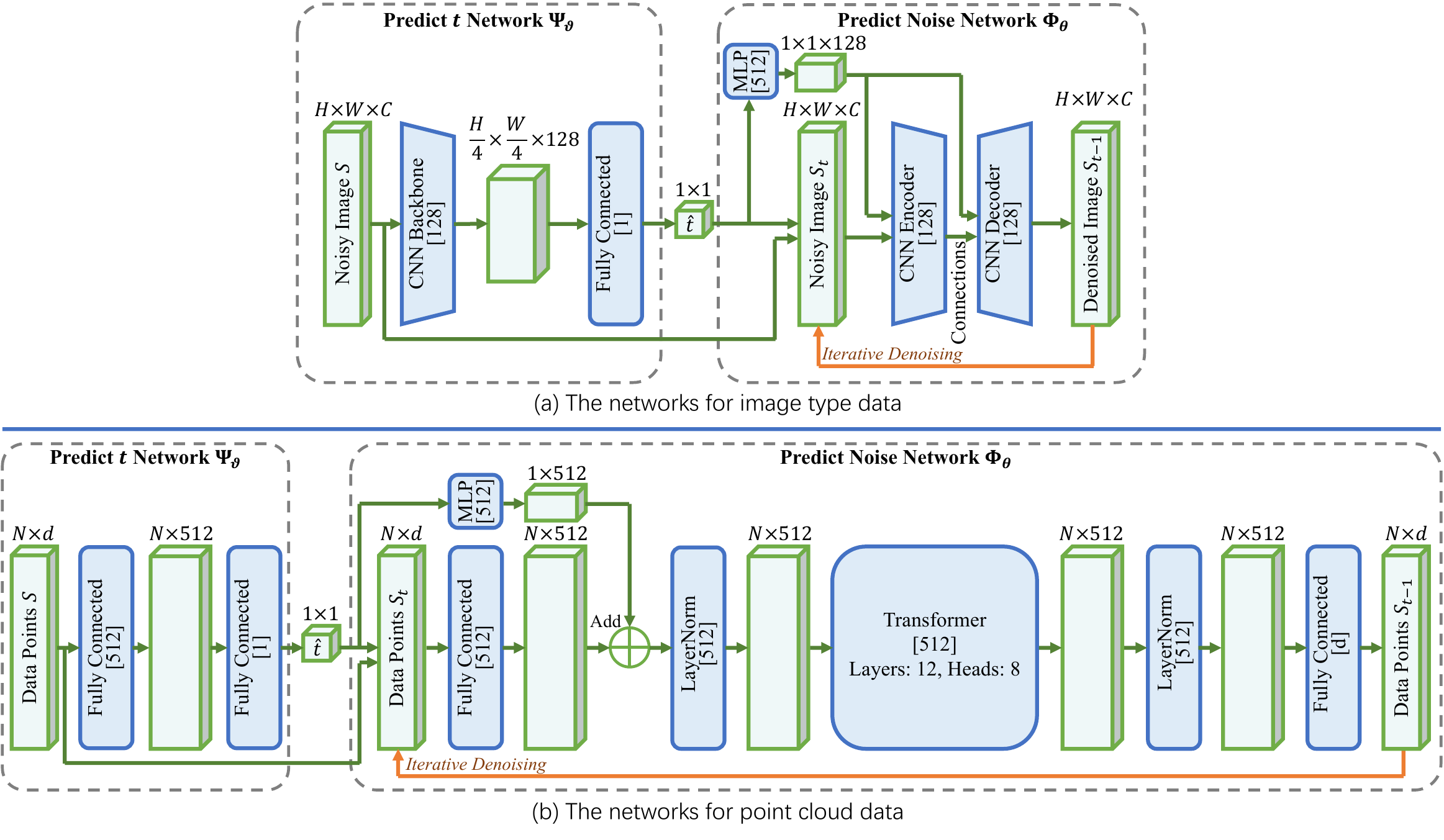}
    \vspace{-0.4cm}
   \caption{\textbf{Neural network architectures for MID.} \textbf{(a)} For image data, a CNN-based architecture is used. The network $\Psi_\vartheta$ (CNN backbone + FC layers) processes the input image to estimate the noise level ($t$). Subsequently, the network $\Phi_\theta$ (CNN encoder-decoder) is applied iteratively, starting from the estimated step $\hat{t}$, to predict and remove noise. \textbf{(b)} For point cloud data and 1D signals, a Transformer-based architecture is employed. The network $\Psi_\vartheta$ (FC layers) estimates the noise level ($t$), and the network $\Phi_\theta$ (Transformer encoder-decoder) performs the iterative noise prediction and removal starting from $\hat{t}$.}
   \label{fig:network}
\end{figure*}

For image data (general images, MRI), MID uses the architecture shown in Fig. \ref{fig:network} (a) for both $\Psi_\vartheta$ and $\Phi_\theta$. The collected noisy image is first processed by a convolutional neural network (CNN) backbone and fully connected layers to predict the current step $\hat{t}$. This noisy image then serves as the initial state $s_t$ in the iterative denoising process. The state $s_t$ is denoised using a CNN encoder and decoder. This process continues until $t=0$, resulting in a clean image.

For point cloud-type data (2D points, correspondences, line segments, amino acid sequences) and 1D signals (sEMG), a Transformer architecture \cite{vaswani2017attention} (encoder-decoder) is utilized. As shown in Fig. \ref{fig:network} (b), the network architecture for these data is similar. Because point cloud is unordered, fully connected layers and a transformer architecture extract features.

All networks are trained end-to-end using the AdamW optimizer with $\beta_1$ = 0.9 and $\beta_2$ = 0.999, an initial learning rate of $1e-4$, and a weight decay of $0.01$. The training uses a global batch size of 8 on an RTX8000 GPU for 150 epochs, saving the final parameters from the last checkpoint.

\section{Experiments}
\label{Experiments}
We first evaluated the performance of MID on several classic computer vision tasks. We then extended the evaluations to bio-related domains to assess its generalizability.

\subsection{Case Study 1: General Image Denoising}
General image denoising is crucial for various computer vision applications. With the increasing prevalence of image-capturing devices, effective denoising is essential to enhance image quality.
For this task, we trained MID using the ILSVRC2012 dataset \cite{deng2009imagenet}, resizing images to $256\times256$ pixels. We evaluated performance on the Kodak dataset \cite{franzen_kodak_1999}, where both Gaussian and Poisson noise were added, following the same preprocessing steps. The denoising quality was assessed using the Peak Signal-to-Noise Ratio (PSNR) and the Structural Similarity Index Measure (SSIM). PSNR measures the pixel-wise intensity difference, while SSIM evaluates similarity in structural information, contrast, and brightness. The formulas for these metrics are as follows:
\begin{equation}
\left\{
\begin{aligned}
\text{PSNR} & = 10 \cdot \log_{10} \left( \frac{\text{MAX}\text{clean}^2}{P_n} \right) \\
\text{SSIM}(I_{\text{clean}}, I_{\text{denoised}}) & = \frac{(2\mu_{\text{clean}}\mu_{\text{denoised}} + C_1)}{(\mu_{\text{clean}}^2 + \mu_{\text{denoised}}^2 + C_1)} \\
& \quad \times \frac{(2\sigma_{\text{clean},\text{denoised}} + C_2)}{(\sigma_{\text{clean}}^2 + \sigma_{\text{denoised}}^2 + C_2)}
\end{aligned}
\right.
\label{eq:PSNR}
\vspace{-0.1cm}
\end{equation}

\begin{figure*}[t]
  \centering
   \includegraphics[width=0.90\linewidth]{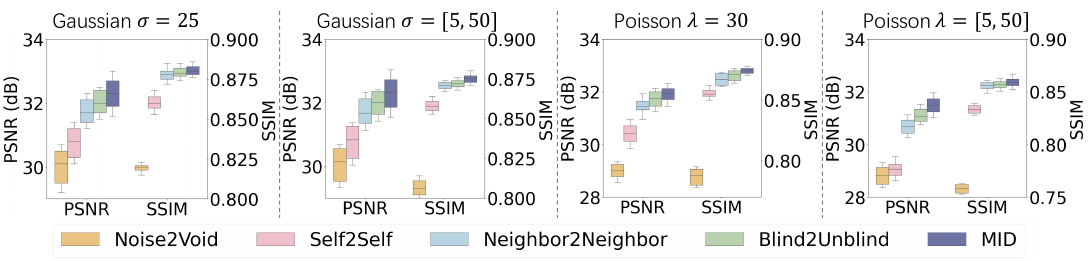}
    \vspace{-0.2cm}
   \caption{\textbf{Quantitative image denoising evaluation on the Kodak dataset.} PSNR and SSIM scores for general image denoising under Gaussian and Poisson noise are presented. MID shows superior performance across both noise types.}
   \label{fig:cv_chart}
\end{figure*}

\begin{figure}[t]
  \centering
   \includegraphics[width=0.98\linewidth]{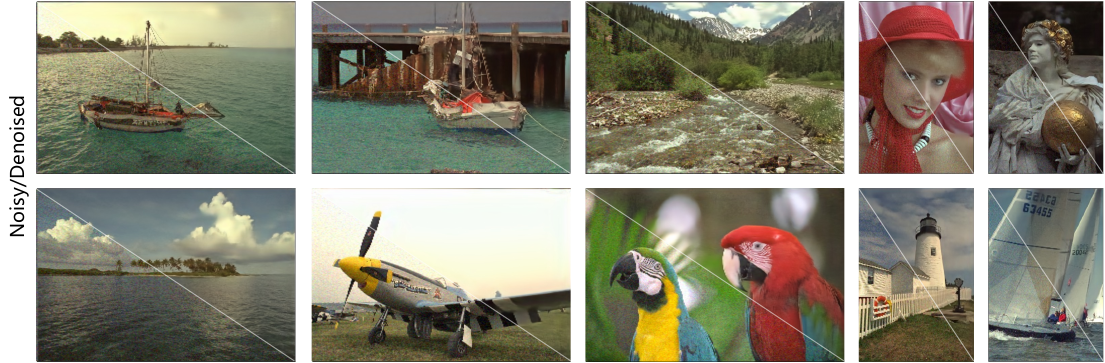}
    \vspace{-0.2cm}
   \caption{\textbf{Visual examples of MID denoising on Kodak images.} MID achieves excellent denoising effects.}
   \label{fig:image}
\end{figure}

Fig. \ref{fig:cv_chart} shows that MID achieves superior PSNR and SSIM performance compared with other SOTA methods in both fixed and variable noise scenarios. While its performance under Poisson noise is slightly lower than that under Gaussian noise, it still surpasses other methods. This demonstrates that MID can handle the challenges posed by non-linear Poisson noise, preserving more image details during iterative denoising. Due to the non-linear properties of Poisson noise, learning its distribution directly is difficult. By employing a first-order Taylor expansion for linearization, MID effectively mitigates this difficulty, enabling accurate denoising. Importantly, because MID was not trained on the Kodak dataset, these results highlight its robust cross-dataset generalization. Qualitative results in Fig. \ref{fig:image} confirm the ability of MID to restore fine textures and details, producing visually coherent images.

\begin{figure*}[t]
  \centering
   \includegraphics[width=0.90\linewidth]{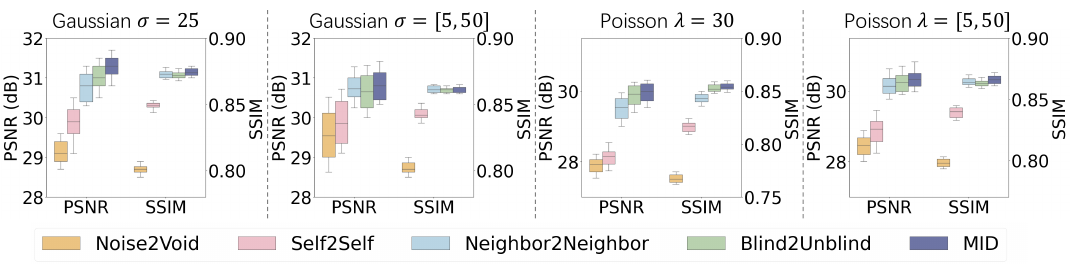}
    \vspace{-0.2cm}
   \caption{\textbf{Denoising performance on the BSD300 dataset.} Quantitative evaluation using Peak Signal-to-Noise Ratio (PSNR) and Structural Similarity Index Measure (SSIM) demonstrates that MID effectively denoises the BSD300 dataset and achieves superior performance compared to other methods.}
   \label{fig:ablation_images}
\end{figure*}

To assess performance, we quantitatively compare MID with other methods using the BSD300 dataset. As shown in Fig. \ref{fig:ablation_images}, MID consistently achieves superior results compared to other methods on BSD300. This suggests that MID effectively adapts to varying noise features, leading to strong denoising performance across different image content. Further qualitative comparisons on both the KODAK and BSD300 datasets, presented in Fig. S1 and Fig. S2 of the Supplementary Material, demonstrate the ability of MID to remove noise and maintain detail under different noise conditions.
\begin{figure}[t]
  \centering
   \includegraphics[width=0.99\linewidth]{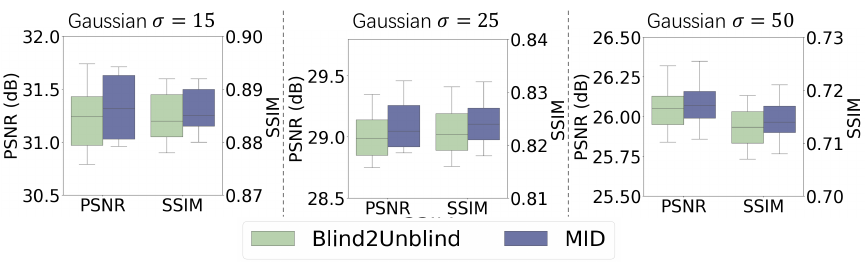}
    \vspace{-0.7cm}
   \caption{\textbf{Denoising performance on BSD68 grayscale images.} This figure presents a quantitative comparison (PSNR and SSIM) of denoising performance for various methods applied to the BSD68 dataset. MID demonstrates significantly superior denoising performance compared to Blind2Unblind.}
   \label{fig:ablation_images_BSD68}
   
\end{figure}

To evaluate performance of MID on single-channel grayscale images, we trained the model on the BSD400 dataset and tested it on the BSD68 dataset. The results in Fig. \ref{fig:ablation_images_BSD68} show that MID outperforms Blind2Unblind, further confirming its general applicability to various image types and its effectiveness in scenarios involving non-linear noise.

\subsection{Case Study 2: Enhancing Robust Estimation}

\begin{figure}[t]
  \centering
   \includegraphics[width=0.8\linewidth]{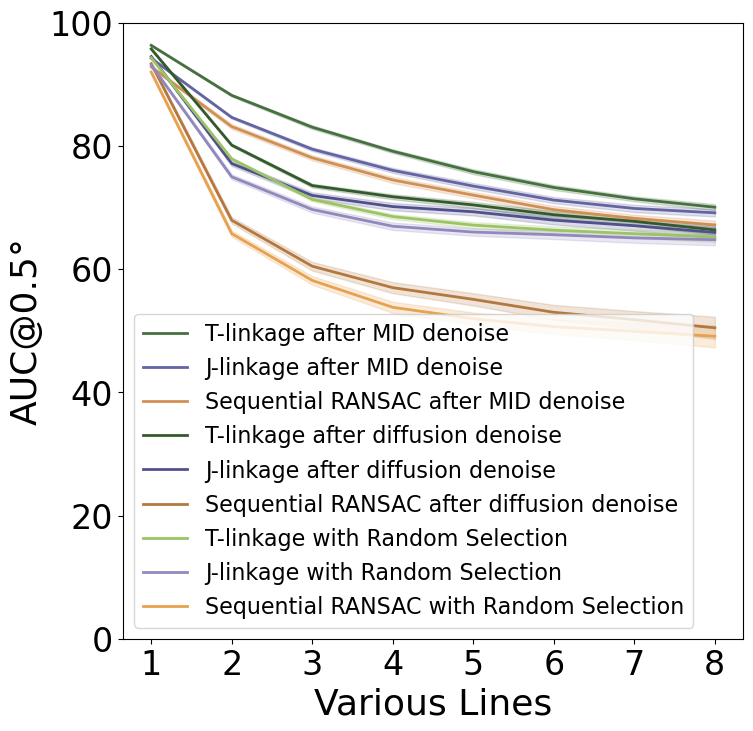}
    \vspace{-0.2cm}
   \caption{\textbf{Area Under the Curve (AUC) for 2D multi-line fitting accuracy.} The comparison of performance on 2D point sets, before and after MID denoising, shows that MID significantly boosts accuracy for various fitting methods (T-linkage, J-linkage, Sequential RANSAC, $p<0.05$).}
   \label{fig:points_chart}
\end{figure}

\begin{figure}[t]
  \centering
   \includegraphics[width=0.99\linewidth]{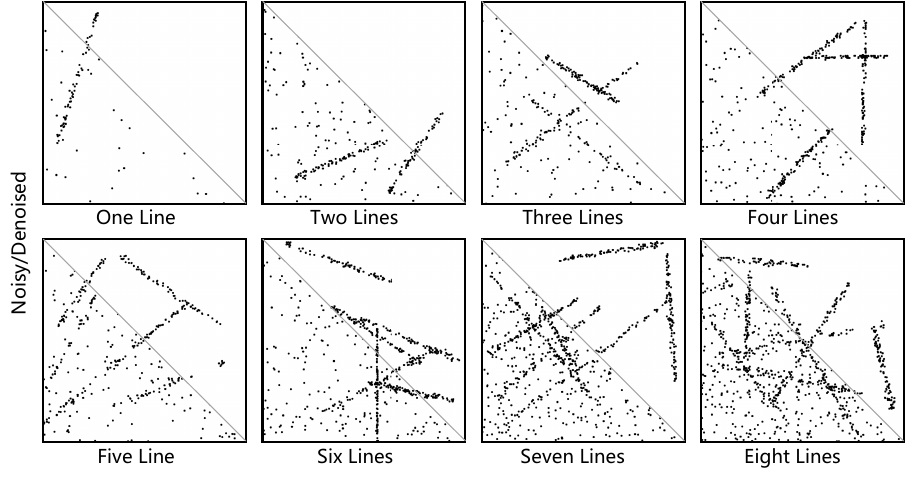}
    \vspace{-0.2cm}
   \caption{\textbf{Visual effect of MID on noisy 2D point sets for multi-line scenes.} MID removes outliers (noise) while preserving the true line structures and their intersections, facilitating more accurate downstream model fitting.}
   \label{fig:points}
\end{figure}

Robust estimation is a critical component in computer vision, playing a crucial role in tasks such as simultaneous localization and mapping (SLAM), 3D reconstruction, and pose estimation. It aims to estimate accurate models from data contaminated with noise and outliers. Traditionally, this challenge is tackled using sampling consensus methods like RANSAC \cite{fischler1981random}, which iteratively sample minimum sets to generate and evaluate hypotheses. However, these approaches can struggle with efficiency and accuracy in highly noisy conditions. We propose that denoising the data before robust estimation can significantly boost the performance of these consensus methods. Therefore, we evaluated the ability of MID to improve solutions for multi-line fitting, fundamental matrix estimation, and vanishing point estimation.

\subsubsection{Improving 2D Multi-Line Fitting}
Fitting multiple lines to noisy 2D point clouds is challenging for methods like Sequential RANSAC \cite{vincent2001detecting}, J-linkage \cite{toldo2008robust}, and T-linkage \cite{magri2014t}. We hypothesize that preprocessing data with MID will improve their accuracy and efficiency.
For this task, we generated a synthetic dataset comprising scenes with 1 to 10 lines (12,000 images per scene count). We used 10,000 images for training and 2,000 for testing. Each image is a $1\times1$ square containing randomly generated lines, with each line segment consisting of 40 to 100 points. We added Gaussian noise (std. dev. 0.007-0.008) and introduced outliers (40\% to 60\% of total points), which are uniformly distributed. Performance was evaluated using the Area Under the Curve (AUC) metric, calculated from a recall curve with a maximum angular error of 0.5 degrees. Recall is defined as:
\begin{equation}
\left\{ \begin{aligned} \text{AUC} &= \sum_{i=1}^{K-1} \left( \frac{\text{Recall}(e_i) + \text{Recall}(e_{i+1})}{2} \right) \cdot (e_{i+1} - e_i) \\ 
\text{Recall} &= \frac{\text{True Positives}}{\text{True Positives} + \text{False Negatives}}
 \end{aligned} \right. .
\label{eq:auc}
\vspace{-0.1cm}
\end{equation}
Here, $e$ represents the angular error between the predicted and ground truth lines.

The results in Fig. \ref{fig:points_chart} demonstrated a significant improvement in line fitting accuracy across all SOTA methods when applied to MID-denoised data. In high-noise, multi-line scenarios, the performance gains reached up to 36.8\% compared to random selection and diffusion-based denoising methods. This highlights MID's effectiveness in denoising point clouds.

Qualitative analysis in Fig. \ref{fig:points} further supports these findings, showing that MID removes outliers while preserving true line structures and crucial intersection points. This enables downstream methods to accurately resolve overlapping models. Fig. S3 and Fig. S4 in the Supplementary Material provide further visualizations, demonstrating that MID effectively removes substantial noise while preserving inlier points, thereby improving the accuracy and efficiency of robust multi-line fitting.

\subsubsection{Strengthening Image Correspondence Matching}
Accurate matching of correspondences between images is essential for many computer vision tasks. These matches, often represented as four-dimensional coordinates, are commonly established by comparing local point descriptors \cite{lowe2004distinctive, arandjelovic2012three}. However, feature-based methods can introduce errors due to feature biases and a lack of global context. These errors negatively impact downstream tasks, such as estimating the fundamental matrix. To mitigate these errors, robust estimation techniques like RANSAC \cite{fischler1981random}, LO-RANSAC \cite{chum2003locally}, and MAGSAC++ \cite{barath2020magsac++} are used. We propose that filtering this noise in advance can be beneficial. We treat sets of image correspondences as multidimensional point clouds, and MID is trained to identify and filter these incorrect matches.

We evaluated the impact of MID on fundamental matrix estimation using the dataset and setup from \cite{barath2020ransac}. Correspondences were detected using RootSIFT \cite{arandjelovic2012three} and matched via nearest neighbor search. The training set included 12 scenes (100,000 image pairs each), and the test set included 2 scenes (4,950 image pairs). We estimated the fundamental matrix from the correspondences and decomposed it to obtain the relative rotation and translation. Performance was measured using the mean average accuracy (mAA) metric \cite{barath2022learning} with a 10-degree tolerance. The accuracy for each scene is the average of an indicator function:
\begin{equation}
mAA = \frac{1}{h} \sum_{i=1}^{h} \left( \frac{1}{k_i} \sum_{j=1}^{k_i} \mathbf{1}(e_{ij} < \epsilon) \right) .
\label{eq:mAA}
\vspace{-0.1cm}
\end{equation}
$e$ is the error, $\epsilon$ is the threshold, $1\left ( \cdot \right ) $ is the indicator function, and $h$ and $k$ are the number of scenes and data points.

\begin{figure}[t]
  \centering
   \includegraphics[width=0.99\linewidth]{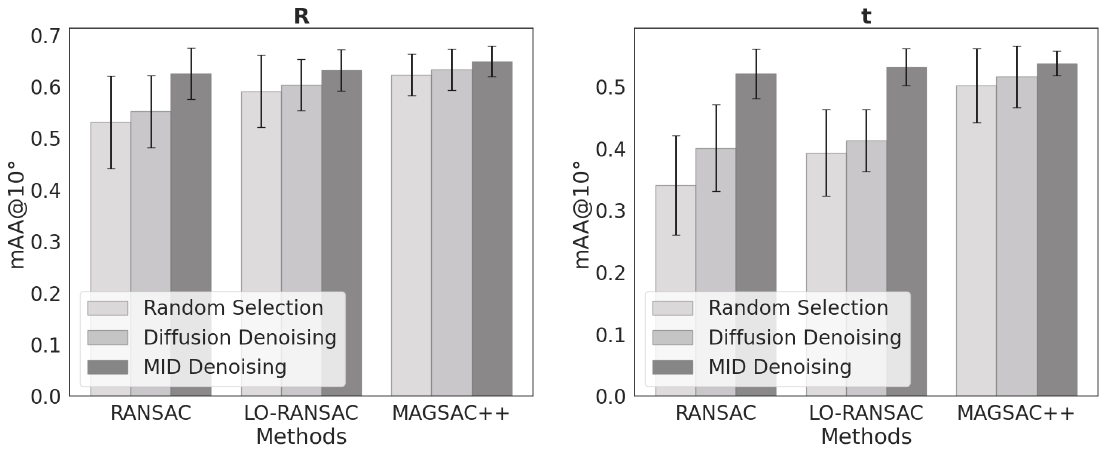}
    \vspace{-0.7cm}
   \caption{\textbf{Mean Average Accuracy (mAA) for relative camera pose estimation from fundamental matrices.} Preprocessing correspondences with MID significantly improves pose accuracy for methods ($p<0.05$).}
   \label{fig:corr_chart}
\end{figure}

\begin{figure}[t]
  \centering
   \includegraphics[width=0.99\linewidth]{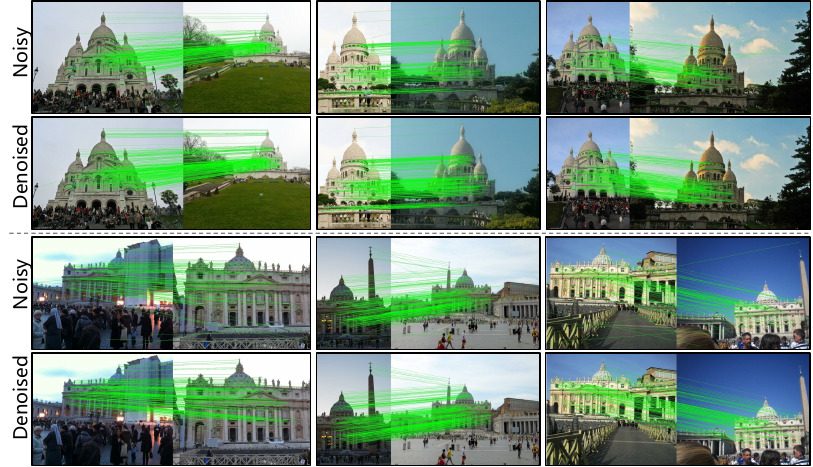}
    \vspace{-0.6cm}
   \caption{\textbf{Qualitative examples of MID denoising image correspondences.} Incorrect matches (noise), in the top row, are filtered by MID, resulting in cleaner correspondences (bottom row) for fundamental matrix estimation.}
   \label{fig:featurepoints}
\end{figure}

The experiments in Fig. \ref{fig:corr_chart} demonstrate that denoising correspondences with MID before fundamental matrix estimation significantly improves the accuracy of the estimated relative camera pose, with improvements up to 53.7\% compared to random selection and diffusion denoising methods. This improvement stems from the ability of MID to model correspondences as point cloud data and learn their underlying features. Unlike methods relying solely on match scores, MID learns broader features of incorrect correspondences, enabling more effective filtering. Fig. \ref{fig:featurepoints} visually illustrates this reduction in incorrect matches. Fig. S5 of the Supplementary Material provides further visual examples across diverse scenes. Furthermore, we investigate the impact of MID-denoised image correspondences for essential matrix estimation.

\begin{figure}[t]
  \centering
   \includegraphics[width=0.99\linewidth]{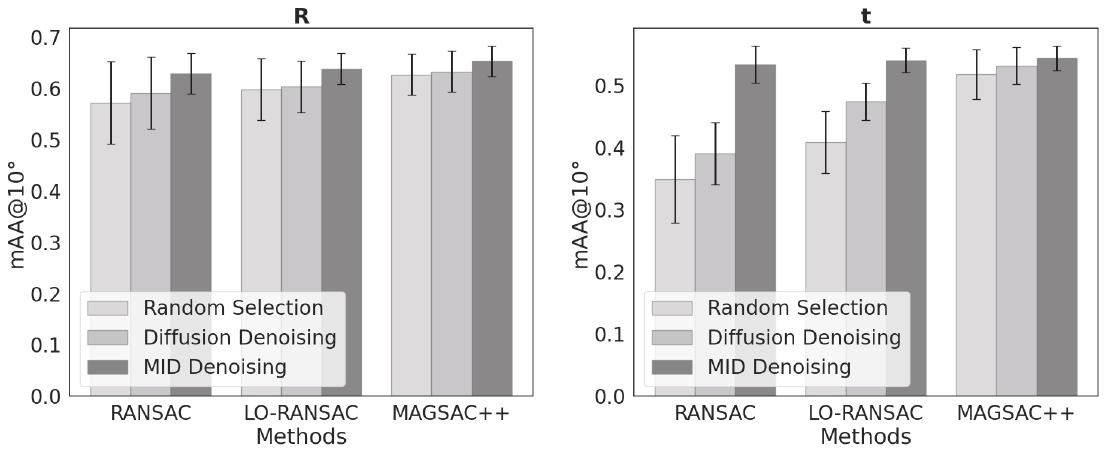}
    \vspace{-0.7cm}
   \caption{\textbf{Performance on essential matrix estimation after MID denoising.} The accuracy of relative camera pose estimation (mAA) derived from the essential matrix is shown. Applying MID to denoise image correspondences improves the accuracy of rotation and translation estimation for all methods.}
   \label{fig:ablation_E}
   
\end{figure}

We further assessed the generalizability of MID by applying it to essential matrix estimation, a task reliant on accurate input correspondences. This task serves as a valuable benchmark for evaluating the effectiveness of MID in correspondence denoising. As shown in Fig. \ref{fig:ablation_E}, MID improves the performance of various essential matrix estimation methods. These findings suggest that MID can effectively denoise a broader range of irregular, multidimensional point cloud data.

\subsubsection{Improving Vanishing Point Estimation}
Estimating vanishing points in single images is crucial for understanding 3D scene geometry, which is essential for applications like autonomous driving and augmented reality. This process typically involves extracting line segments and inferring vanishing points from their orientations. However, many extracted line segments are noisy and do not accurately represent the scene geometry, hindering precise estimation. Methods like J-Linkage \cite{toldo2008robust}, T-Linkage \cite{magri2014t}, CONSAC \cite{kluger2020consac}, and PARSAC \cite{kluger2024parsac} attempt to handle this.

We treat the extracted line segments as an unordered, multidimensional point cloud. The goal is to remove noisy line segments before vanishing point estimation. MID is trained by iteratively adding random line segments as noise, enabling it to learn data features and effectively denoise the input.

\begin{figure}[t]
  \centering
   \includegraphics[width=0.99\linewidth]{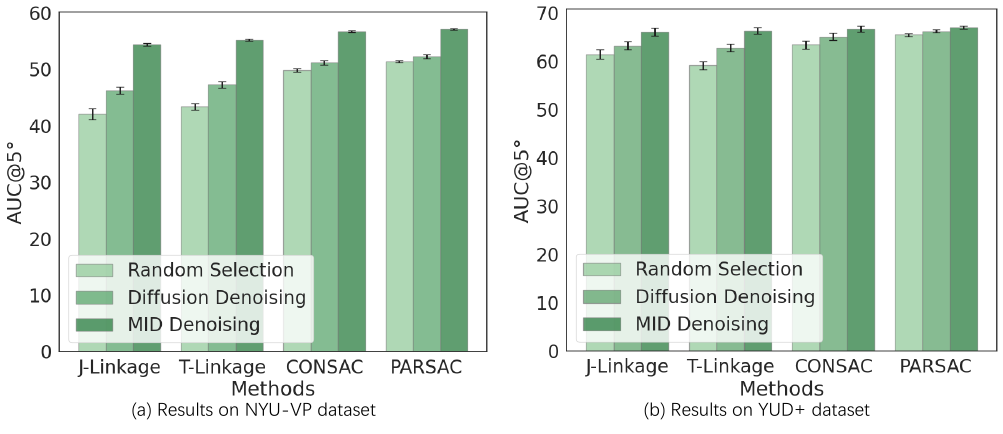}
    \vspace{-0.6cm}
   \caption{\textbf{Quantitative evaluation of vanishing point estimation.} (a) The mAA for vanishing point estimation on line segments from the NYU-VP dataset. (b) The mAA for vanishing point estimation on line segments from the YUD+ dataset. MID preprocessing significantly improves performance.}
   \label{fig:vp_chart}
\end{figure}

\begin{figure*}[t]
  \centering
   \includegraphics[width=0.99\linewidth]{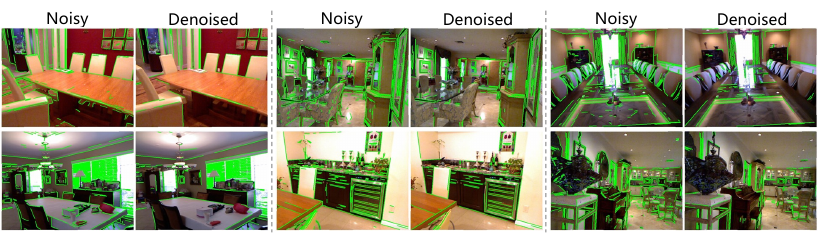}
    \vspace{-0.3cm}
   \caption{\textbf{Visual comparison of line segments for vanishing point estimation.} The original noisy line segments (left column) are filtered by MID (right column), which effectively removes extraneous segments while preserving those relevant to the scene's geometric structure.}
   \label{fig:vp}
\end{figure*}

To quantitatively evaluate the impact of MID, we trained and tested several algorithms on the NYU-VP dataset \cite{kluger2020consac} (1,224 training scenes, 225 test scenes), which includes line segments extracted via LSD \cite{vincent2001detecting}. Performance was quantified using the AUC metric (Eq. \ref{eq:auc}) with a 5-degree threshold, as described for the 2D point denoising task. Here, the error used for AUC calculation is the angular error between the estimated and ground truth vanishing points.

The quantitative results in Fig. \ref{fig:vp_chart} (a) show that using MID to denoise line segments significantly improves performance for each method compared to random selection and diffusion denoising. Furthermore, the reduced standard deviation indicates improved stability of the downstream methods after MID noise removal. Qualitative results in Fig. \ref{fig:vp} visually demonstrate that MID effectively removes invalid line segments while preserving those relevant for vanishing point estimation. Further visualizations are in Fig. S6 of the Supplementary Material illustrate the impact of MID across various scenarios. To assess cross-dataset generalization, we evaluated performance on the YUD+ dataset \cite{kluger2020consac}. As shown in Fig. \ref{fig:vp_chart} (b), despite not being trained on YUD+, MID effectively improved the performance of various methods, suggesting it learns generalizable features of noisy line segments. These experiments demonstrate that MID can handle irregular and non-coordinate data points, highlighting its strong generalization capabilities.

\subsection{Case Study 3: Denoising Biological Signals}
Biosignal monitoring is essential for managing health and diagnosing medical conditions. Surface electromyography (sEMG), a non-invasive technique capturing muscle activity, is widely used in medical diagnosis, rehabilitation, and assistive technologies. However, sEMG recordings are often contaminated by electrocardiogram (ECG) signals.

To evaluate the denoising capabilities of MID on biological signals, we used sEMG and ECG data from the NINAPro \cite{atzori2014electromyography} and PhysioNet databases \cite{goldberger2000physiobank}, respectively. The sEMG signals from the DB2 subset of the Non-Invasive Adaptive Prosthetics (NINAPro) database form the basis for this evaluation. This subset includes 12 channels of sEMG recordings of hand movements from 40 subjects. This subset includes data from Exercises 1, 2, and 3, which involve 17, 22, and 10 movements, respectively. Each movement is repeated six times for five seconds, with three-second rest intervals. Electrocardiogram (ECG) interference is simulated using data from the MIT-BIH Normal Sinus Rhythm Database (NSRD) from PhysioNet, which contains two-channel ECG recordings from 18 healthy subjects sampled at 128 Hz. Signal reconstruction quality is assessed using four metrics. 

For training and validation, we used sEMG segments from Channel 2 (Exercises 1 and 3) from 30 subjects, corrupted with ECG signals from 12 NSRD subjects at six signal-to-noise ratios (SNRs): -5 to -15 dB. For validation, we used ECG signals from three different NSRD subjects at the same SNRs. This approach uses actual recorded signals, providing a more accurate assessment of performance in real-world scenarios.

To assess generalizability, we created a test set with deliberately mismatched conditions: sEMG segments from Channels 9-12 (Exercise 2) from the remaining 10 subjects, corrupted with ECG data from three other NSRD subjects at SNRs from -12 to -6 dB. The sEMG subjects, movements, channels, ECG subjects, and SNRs in the test set were completely distinct from the training data.

Signal reconstruction quality is assessed using four metrics. The Signal-to-Noise Ratio (SNR) improvement ($SNR_{imp}$) quantifies the change in SNR after denoising:
\begin{equation}
SNR_{imp} = 10 \cdot \log_{10} \left( \frac{P_{noisy}}{P_{noise}} \right) - 10 \cdot \log_{10} \left( \frac{P_{denoise}}{P_{noise}} \right) ,
\label{eq:SNRimp}
\vspace{-0.1cm}
\end{equation}
where the power terms are calculated as:
\begin{equation}
\left\{ \begin{aligned} P_{noisy} = \frac{1}{L} \sum_{n=1}^{L} \left| \text{Signal}_{noisy} [n] \right|^2 \\ 
P_{denoise} = \frac{1}{L} \sum_{n=1}^{L} \left| \text{Signal}_{denoise}[n] \right|^2\\
P_{noise} = \frac{1}{L} \sum_{n=1}^{L} \left| \text{Signal}_{noise}[n] \right|^2
 \end{aligned} \right. .
\vspace{-0.1cm}
\end{equation}
Here, the $L$ is the signal length. A larger $SNR_{imp}$ indicates better noise reduction.

Root Mean Squared Error (RMSE) measures the difference between the denoised and clean signals:
\begin{equation}
\text{RMSE} = \sqrt{\frac{1}{L} \sum_{n=1}^{L} \left( \text{Signal}_{clean}[n] - \text{Signal}_{denoise}[n] \right)^2} .
\vspace{-0.1cm}
\end{equation}
Lower RMSE values indicate better denoising. Feature extraction accuracy is evaluated using RMSE of the average rectified value (ARV) and RMSE of the mean frequency (MF):
\begin{equation}
\left\{ \begin{aligned} &\text{RMSE}_{ARV} = \sqrt{\frac{1}{L} \sum_{i=1}^L (ARV_{\text{clean}} - ARV_{\text{denoised}})^2} \\ 
&\text{RMSE}_{MF} = \sqrt{\frac{1}{L} \sum_{i=1}^L (MF_{\text{clean}} - MF_{\text{denoised}})^2}
 \end{aligned} \right. ,
\label{eq:RMSEARV}
\vspace{-0.1cm}
\end{equation}
where ARV and MF are calculated as:
\begin{equation}
\left\{ \begin{aligned} \text{ARV} &= \frac{1}{L} \sum_{i=1}^L |\text{Signal}_i| \\ 
\text{MF} &= \frac{\sum_{i=1}^L f_i \cdot P_i}{\sum_{i=1}^L P_i}
 \end{aligned} \right. .
\label{eq:ARV}
\vspace{-0.1cm}
\end{equation}
The $f_i$ and $P_i$ are the frequency components and power spectral density, respectively. Lower RMSE values for ARV and MF indicate better denoising.

\begin{figure*}[t]
  \centering
   \includegraphics[width=0.90\linewidth]{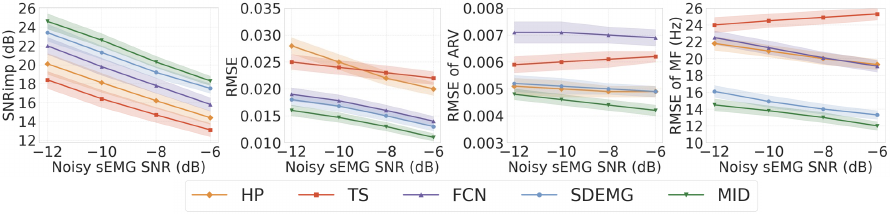}
    \vspace{-0.2cm}
   \caption{\textbf{Quantitative evaluation of sEMG denoising.} Performance metrics ($SNR_{imp}$, RMSE, ARV, MF) across varying ECG interference levels are shown. MID consistently outperforms other methods (high-pass filter, template subtraction, FCN, SDEMG) ($p<0.05$).}
   \label{fig:emg_chart}
\end{figure*}

Fig. \ref{fig:emg_chart} shows MID outperforming SOTA methods including high-pass filtering, template subtraction, a standard Fully Convolutional Network (FCN), and SDEMG ($p<0.05$). Importantly, MID maintains stable performance across all tested SNRs, unlike other methods. This robustness likely stems from the iterative optimization of signal details inherent in MID, providing superior denoising compared to non-iterative approaches. These findings suggest strong potential for clinical applications of MID in sEMG denoising.

\begin{figure}[t]
  \centering
   \includegraphics[width=0.99\linewidth]{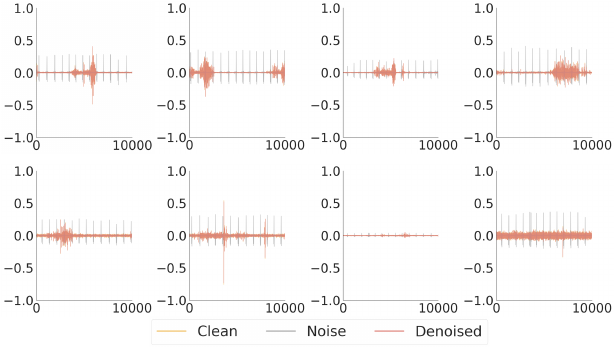}
    \vspace{-0.4cm}
   \caption{\textbf{Visual comparison of sEMG signals.} The original clean signal, noisy signal (sEMG + ECG), and MID-denoised signal are shown. MID effectively removes ECG interference while preserving sEMG characteristics.}
   \label{fig:emg}
\end{figure}

Qualitative results of Fig. \ref{fig:emg} further illustrate the effectiveness of MID. It accurately removes ECG noise, recovering sEMG signals that closely resemble the original clean signals. MID preserves important signal details and significantly suppresses ECG interference. This indicates that MID can effectively capture the features of both sEMG and ECG signals through self-supervised learning, without requiring clean ground truth data. Additional visualizations of waveform comparisons in Fig. S7 of the Supplementary Material demonstrate the strong denoising performance of MID on sEMG signals, further validating its effectiveness across diverse signal types.

These experiments demonstrate that MID has strong generalization abilities, extending beyond specific noise distributions to handle complex noise in real-world recordings. Moreover, these results show that MID can process one-dimensional signal data, opening possibilities for its use in other signal processing applications.

\subsection{Case Study 4: Denoising Medical Data}
Advancements in medical imaging technology have led to the widespread use of medical images for monitoring health conditions and diagnosing diseases. Magnetic Resonance Imaging (MRI) is a key clinical tool providing detailed anatomical and functional insights. However, clinical practice often prioritizes shorter scan times, which can lead to lower signal-to-noise ratios (SNR). We applied MID to MRI denoising. MRI scans are represented as a 4D sequence $X\in \mathbb{R} ^{w \times h \times d \times l}$, where $w \times h$ denotes the axial 2D slices. Following DDM$^2$ \cite{xiangddm}, MID is trained by iteratively adding Gaussian noise to MRI slices from a training set, enabling it to learn the specific noise characteristics in this modality. We evaluated MID on the Stanford HARDI dataset \cite{rokem2016stanford}, comparing it against state-of-the-art (SOTA) methods.

We evaluated MID and state-of-the-art (SOTA) algorithms on the Stanford HARDI dataset \cite{rokem2016stanford} and Sherbrooke 3-shell dataset \cite{garyfallidis2014dipy} for training and evaluation. The HARDI dataset comprises 4D sequences with dimensions $106\times81\times76\times150$ and a b-value of 2000. The Sherbrooke 3-shell dataset \cite{garyfallidis2014dipy} is a $128\times128 \times 64 \times 193$ dataset with b-value=1000. Before processing by the neural network, axial two-dimensional slices were resized to $256\times256$ pixels and normalized to the range $\left [ -1,1 \right ]$. A key challenge in evaluating MRI denoising performance is the lack of corresponding clean data in these datasets. To overcome this, we adopted the approach used in a previous study with DDM$^2$ and use Signal-to-Noise Ratio (SNR) and Contrast-to-Noise Ratio (CNR) as proxy metrics:
\begin{equation}
\left\{ \begin{aligned} \text{SNR} & = 10 \log_{10} \left( \frac{P_s}{P_n} \right) \\ \text{CNR} & = \frac{\overline{I_{denoised}} - \overline{I_{noisy}}}{\sigma_{noisy}} \end{aligned} \right. ,
\label{eq:SNR}
\vspace{-0.1cm}
\end{equation}
where $P_s$ and $P_n$ represent the power of the denoised signal and the noise, respectively, calculated as:
\begin{equation}
\left\{ \begin{aligned} P_s &= \frac{1}{WH} \sum_{i=1}^{W} \sum_{j=1}^{H} I_{\text{denoised}}(i,j)^2 \\ P_n &= \frac{1}{WH} \sum_{i=1}^{W} \sum_{j=1}^{H} (I_{\text{noisy}}(i,j) - I_{\text{denoised}}(i,j))^2 \end{aligned} \right. .
\label{eq:power}
\vspace{-0.1cm}
\end{equation}
Here, $W$ and $H$ are image width and height, $I$ is the image, and $\sigma$ is the standard deviation.

\begin{figure}[t]
  \centering
   \includegraphics[width=0.75\linewidth]{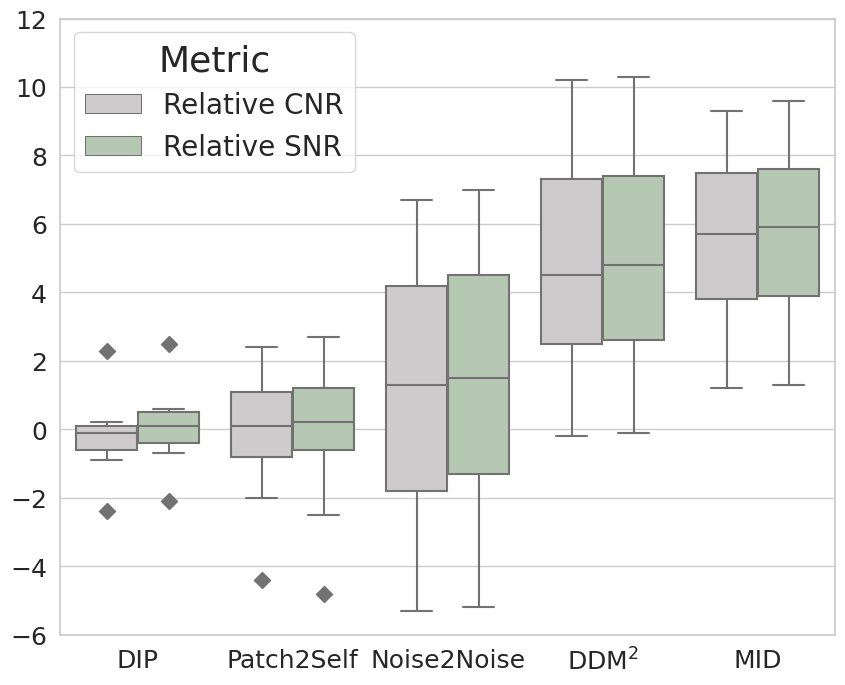}
    \vspace{-0.2cm}
   \caption{\textbf{Quantitative MRI denoising performance comparison.} Relative Signal-to-Noise Ratio (SNR) and Contrast-to-Noise Ratio (CNR) for MID and other methods. Box plots display median (center line), interquartile range (box), and 1.5x IQR (whiskers). MID demonstrates superior performance.}
   \label{fig:mri_chart}
\end{figure}

Quantitative results in Fig. \ref{fig:mri_chart} show MID significantly outperforming DDM$^2$, Noise2Noise, Patch2Self, and Deep Image Prior (MIDP) in both relative SNR and relative CNR ($p<0.05$). Notably, MID achieved these results using data from only a single volume for denoising a given target, unlike some methods that benefit from inter-volume redundancies. This improvement is attributed to the comprehensive learning of MID across various noise levels and its iterative refinement, which preserves crucial image details.

\begin{figure}[t]
  \centering
   \includegraphics[width=0.85\linewidth]{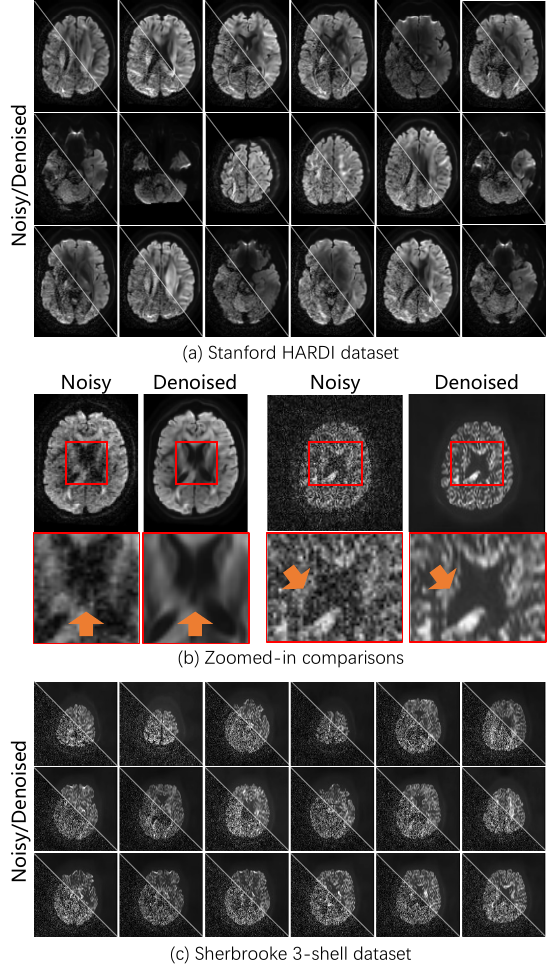}
    \vspace{-0.2cm}
   \caption{\textbf{Visual examples of MRI denoising.} \textbf{(a)} MID denoising on the Stanford HARDI dataset shows effective noise removal while preserving anatomical details. \textbf{(b)} Zoomed-in comparisons highlight the ability of MID to recover fine edge details (arrows). \textbf{(c)} Denoising performance on the Sherbrooke 3-shell dataset illustrates generalization capability.}
   \label{fig:mri}
\end{figure}

Qualitative results in Fig. \ref{fig:mri} (a) and (b) show that MID demonstrates superior noise suppression while effectively recovering and enhancing anatomical details. The framework's ability to distinguish signal from noise, learned through its adding noise process, is evident. Extensive visualizations in Fig. S8 of the Supplementary Material support these findings and confirm the superior performance of MID.

To assess robustness, MID was also tested on the Sherbrooke 3-shell dataset \cite{garyfallidis2014dipy}, shown in Fig. \ref{fig:mri} (c). The results show that MID generalizes well to different datasets, maintaining high SNR and CNR. Additional visualizations in Fig. S9 of the Supplementary Material further confirm this robust performance, highlighting MID's potential for improving diagnostic confidence in clinical MRI.

\subsection{Case Study 5: Strengthening Amino Acid Sequences in bioinformatics}
In biology, accurate protein contact prediction is essential for precise protein structure prediction, and Multiple amino acid Sequence Alignment (MSA) significantly improves this accuracy \cite{rao2021msa}. Unlike image and signal data, an MSA consists of discrete and unstructured sequences. We treat it as multidimensional point cloud data. Large MSAs can be computationally intensive, making it crucial to select high-quality sequences.

We reframe the task of selecting informative sequences from an MSA as a denoising problem. Each amino acid sequence is treated as a data point in a high-dimensional discrete space. Large MSAs often contain highly redundant sequences (i.e., those with low average Hamming distances to others), which increase computational cost while adding little information. These redundant sequences are therefore treated as ``noise" in this context. MID is then trained by iteratively adding these noisy sequences into subsets of the MSA, learning to distinguish more informative sequences.

\begin{figure*}[t]
  \centering
   \includegraphics[width=0.75\linewidth]{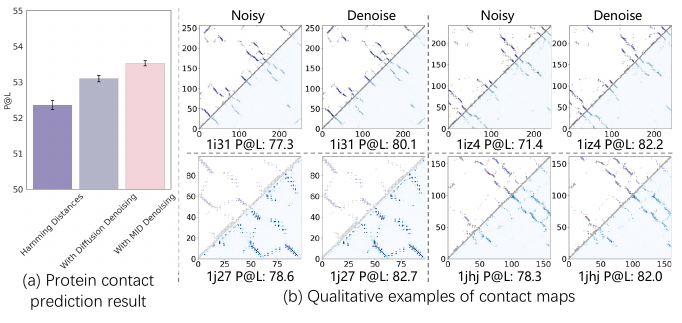}
    \vspace{-0.4cm}
   \caption{\textbf{Evaluation of MID for amino acid sequence denoising.} \textbf{(a)} Impact of MID-denoised sequences on downstream protein contact prediction (MSA Transformer), measured by Top-L long-range contact precision (P@L). MID preprocessing improves prediction accuracy. \textbf{(b)} Visual examples of contact maps for different proteins, showing improved contact prediction (higher P@L scores) after MID denoising.}
   \label{fig:msa_chart}
\end{figure*}

For this task, we used the UniClust30 dataset \cite{mirdita2017uniclust} for training and evaluation. Protein sequences were encoded using the token module of the MSA Transformer. Denoising performance was assessed by predicting contacts on the denoised sequences and measuring Top-L long-range contact precision. This metric is chosen because long-range contact prediction is more challenging than short-range prediction and is thus more sensitive to the effects of denoising. This metric also aligns with the practical application of capturing 3D protein structural information.
As shown in Fig. \ref{fig:msa_chart} (a), contact predictions using MID-processed sequences yielded a significant improvement in Top-L long-range contact precision (P@L) by 2.2\% on average compared to the standard sequence selection method and a diffusion denoising method ($p<0.05$). This indicates the ability of MID to identify and retain higher-quality sequences that are more conducive to accurate contact prediction. 

We further quantified the impact of MID on long-range contact prediction using the Top-L long-range contact precision metric. The results in Fig. \ref{fig:msa_chart} (b) consistently demonstrate that MID enhances contact prediction accuracy, especially for long-range contacts. Notably, the ``noise" in this task consists of real amino acid sequences, showing that MID can evaluate data quality beyond simple noise distributions. Further examples of MID processing different protein sequences are shown in Fig. S10 of the Supplementary Material. These findings highlight MID's potential for various biological applications, including protein structure prediction.

\subsection{Ablation Study}
To assess the importance of iterative noise addition in MID, we conducted further experiments on both Poisson noisy image denoising and correspondence denoising. We compared our iterative approach with a one-shot denoising method, where the model attempts to remove noise in a single step.

\begin{figure}[t]
  \centering
   \includegraphics[width=0.99\linewidth]{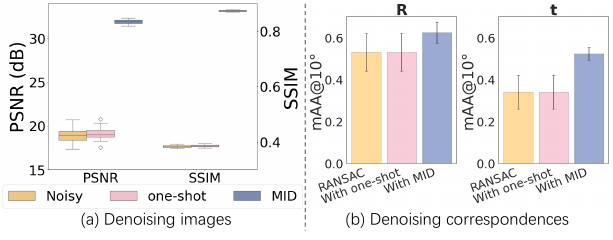}
    \vspace{-0.7cm}
   \caption{\textbf{Comparison of iterative vs. one-shot denoising.} \textbf{(a)} Performance on images corrupted by Poisson noise. The one-shot approach is ineffective in removing complex, non-linear Poisson noise. \textbf{(b)} Performance on denoising image correspondences (point cloud data). The one-shot approach fails to denoise the correspondences.}
   \label{fig:ablation_chart}
\end{figure}

\begin{figure}[t]
  \centering
   \includegraphics[width=0.99\linewidth]{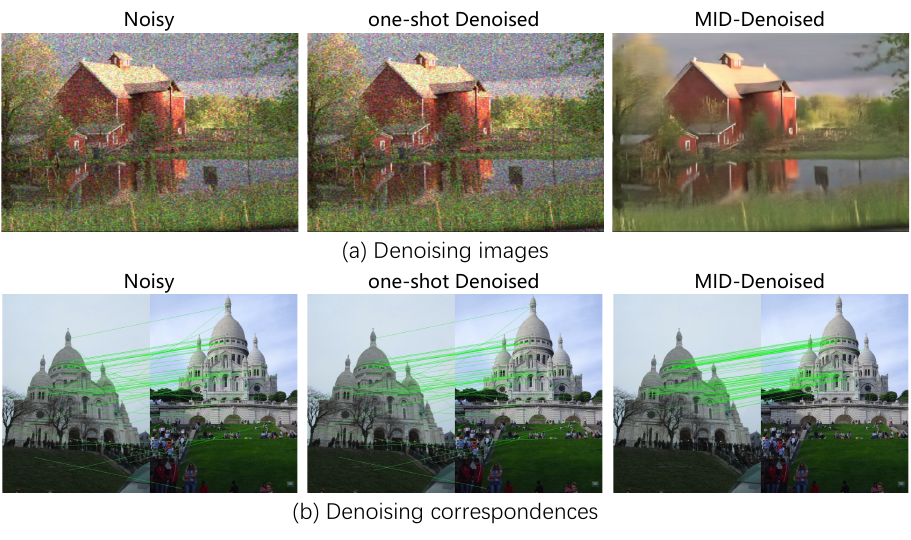}
    \vspace{-0.7cm}
   \caption{\textbf{Qualitative evaluation of the iterative denoising effect.} \textbf{(a)} Visual results confirm the inadequacy of the one-shot method for image denoising. \textbf{(b)} Visual results further demonstrate the failure of the one-shot approach to denoise image correspondences. These findings validate the necessity of MID's iterative approach.}
   \label{fig:ablation}
\end{figure}

The quantitative results in Fig. \ref{fig:ablation_chart} (a) and (b) demonstrate that one-shot denoising is ineffective for both types of data. The data remains largely unchanged after one-shot denoising, resulting in similar performance metrics before and after the process. This failure is attributed to the absence of a Taylor expansion to linearize the noise removal process. Neural networks struggle to directly learn the large differences between noisy and clean data in a single step. The Taylor expansion transforms a complex, non-linear process into a multi-step linearization. In this iterative process, the neural network only needs to learn the data changes at each step, simplifying the learning task and facilitating convergence. The qualitative results in Fig. \ref{fig:ablation} (a) and (b) support this conclusion.

\section{Discussion}
\label{discussion}

The methodology of MID, involving the progressive addition and removal of noise, bears a procedural resemblance to denoising diffusion probabilistic models (DDPMs). However, fundamental differences exist in their objective, supervision, and applicability. Regarding their training objective, diffusion models are trained to reconstruct clean data from its noisy version at every step, minimizing a noise prediction loss derived from a fixed forward diffusion schedule. In contrast, MID learns to recognize and subtract injected noise increments, guided by a step prediction network, without assuming a fixed forward process. Their data requirements also differ significantly. Diffusion models typically need large datasets of clean examples for training. MID, however, operates in a self-supervised manner, learning directly from noisy-only datasets without requiring clean–noisy pairs. This distinction is critical: MID is not a generative model but an iterative, self-supervised noise subtractor that directly learns the structure of noise present in the target domain.

In addition, MID offers several distinct advantages over traditional and other learning-based denoising approaches. A key strength is its cross-modal generality; the same core architecture operates effectively on images, 1D signals, point clouds, and discrete sequences with minimal adaptation. This flexibility stems from its modular two-network design, which decouples step estimation from noise prediction, allowing the framework to adapt across modalities without relying on domain-specific assumptions. Furthermore, MID's iterative refinement strategy is particularly effective for preserving fine details. Instead of attempting to remove all noise in a single pass, which often results in oversmoothing, MID performs multiple, smaller corrections. This enables the model to maintain edges, textures, and subtle features. Finally, the framework adeptly handles non-linear noise by employing a first-order Taylor expansion. This local linearization of complex contamination processes allows MID to tackle structured, signal-dependent noise patterns that confound purely linear denoisers, enhancing its utility in real-world scenarios.

\section{Limitations and Future Work}
\label{Limitations}

While MID demonstrates robust performance across diverse domains, several limitations present opportunities for future work. The framework's reliance on iterative subtraction may struggle in extreme noise scenarios where signal structures are almost entirely obscured, as it depends on partial structural cues. Future research could explore integrating generative priors to enhance reconstruction in such low-signal regimes. Additionally, the iterative refinement strategy, while crucial for detail preservation, inherently increases inference time compared to single-pass denoisers. This latency could be a bottleneck for real-time applications. Consequently, future work will focus on model optimization and acceleration, investigating techniques such as knowledge distillation or developing more efficient sampling strategies to reduce the number of required denoising steps without sacrificing performance.

\section{Conclusion}
We presented MID, a self-supervised, Multimodal Iterative Denoising framework. MID models noise as a state within a continuous process of non-linear noise accumulation. By combining a two-network design (step prediction and noise prediction) with a first-order Taylor expansion for non-linear contamination, this approach enables robust, detail-preserving restoration without needing paired clean–noisy datasets. Extensive experiments on computer vision, biological signals, medical imaging, and bioinformatics confirm the cross-modal versatility and competitive performance of MID against specialized state-of-the-art methods. In summary, MID provides a unified, self-supervised approach to denoising across diverse domains and noise types. Its iterative refinement enables robust and generalizable performance without requiring paired data. This work opens new possibilities for scalable noise removal in scientific and engineering fields.



 
%

\normalem
\bibliographystyle{IEEEtran}
\bibliography{main}

\vfill

\end{document}